%% file: root.tex
\documentclass[letterpaper, 10 pt, journal, twoside]{IEEEtran} % final submission

\ifCLASSINFOpdf
\else
\fi

\usepackage{graphics} % for pdf, bitmapped graphics files
\usepackage{epsfig} % for postscript graphics files
\usepackage{mathptmx} % assumes new font selection scheme installed
\usepackage{times} % assumes new font selection scheme installed
\usepackage{amsmath} % assumes amsmath package installed
\usepackage{amssymb}  % assumes amsmath package installed

\usepackage{multirow}
\usepackage[table,xcdraw]{xcolor}
\usepackage{threeparttable}
\usepackage{eucal}
\usepackage{ulem}
\usepackage{balance}
\usepackage{mathrsfs}  
\usepackage{comment}
\usepackage{algorithm}
\usepackage[noend]{algpseudocode}
\usepackage{float}
\usepackage{textcomp}
\usepackage{mathcomp}
\usepackage{bm}
\usepackage{url}

\makeatletter
\let\MYcaption\@makecaption
\makeatother

\usepackage[font=footnotesize]{subcaption}

\makeatletter
\let\@makecaption\MYcaption
\makeatother

\usepackage{subcaption}

\algnewcommand{\Break}{\textbf{break}}

\newcommand{\argmax}{\mathop{\rm argmax}\limits}
\newcommand{\argmin}{\mathop{\rm argmin}\limits}

\newcommand{\myvec}[1]{\mathbf{#1}}
\newcommand{\myarray}[1]{\textbf{\textit{#1}}}
\newcommand{\myset}[1]{\mathit{#1}}

\usepackage[colorinlistoftodos]{todonotes}

\begin{document}

\title{
Generalized LOAM: LiDAR Odometry Estimation with Trainable Local Geometric Features
}

\author{Kohei Honda$^{1}$, Kenji Koide$^{2}$, Masashi Yokozuka$^{2}$, Shuji Oishi$^{2}$, and Atsuhiko Banno$^{2}$% <-this % stops a space
\thanks{Manuscript received: June, 10, 2022; Revised September, 21, 2021; Accepted October, 17, 2021.}%Use only for final RAL version
\thanks{This paper was recommended for publication by Editor Javier Civera upon evaluation of the Associate Editor and Reviewers' comments.} %Use only for final RAL version
\thanks{This work was supported in part by the DII Collaborative Graduate Program commissioned by Nagoya University, a project commissioned by the New Energy and Industrial Technology Development Organization (NEDO), and JSPS KAKENHI (Grant Number 22K12214).}% <-this % stops a space
\thanks{$^{1}$Kohei Honda is with the Department of Mechanical Systems Engineering, Graduate School of Engineering, Nagoya University, Furo-cho, Chikusa-ku, Nagoya, Aichi, Japan, {\tt\small honda.kohei.b0@s.mail.nagoya-u.ac.jp}}%
\thanks{$^{2}$Kenji Koide, Masashi Yokozuka, Shuji Oishi, and Atsuhiko Banno are with the Department of Information Technology and Human Factors, the National Institute of Advanced Industrial Science and Technology, Umezono 1-1-1, Tsukuba, 3050061, Ibaraki, Japan}%
% , {\tt\small \{k.koide, yokotsuka-masashi, shuji.oishi, atsuhiko.banno \}@aist.go.jp}}%
\thanks{Digital Object Identifier (DOI): see top of this page.}
}

\markboth{IEEE ROBOTICS AND AUTOMATION LETTERS. PREPRINT VERSION. ACCEPTED OCTOBER, 2022}
{Honda \MakeLowercase{\textit{et al.}}: Generalized LOAM: LiDAR Odometry Estimation with Trainable Local Geometric Features} 

% For arxive preprint
\copyright 2022 IEEE.  Personal use of this material is permitted. 
Permission from IEEE must be obtained for all other uses, in any current or future media, including reprinting/republishing this material for advertising or promotional purposes, creating new collective works, for resale or redistribution to servers or lists, or reuse of any copyrighted component of this work in other works.

\newpage

\maketitle
% \thispagestyle{empty}
% \pagestyle{empty}

%%%%%%%%%%%%%%%%%%%%%%%%%%%%%%%%%%%%%%%%%%%%%%%%%%%%%%%%%%%%%%%%%%%%%%%%%%%%%%%%
\begin{abstract}
\input{abstract}

\end{abstract}


\IEEEpeerreviewmaketitle

%%%%%%%%%%%%%%%%%%%%%%%%%%%%%%%%%%%%%%%%%%%%%%%%%%%%%%%%%%%%%%%%%%%%%%%%%%%%%%%%

% MAIN PARTS
\input{introduction}

\input{related_works}
\input{GeneralizedLOAM}

\input{experiments}
\input{conclusion}

%%%%%%%%%%%%%%%%%%%%%%%%%%%%%%%%%%%%%%%%%%%%%%%%%%%%%%%%%%%%%%%%%%%%%%%%%%%%%%%%

%%%%%%%%%%%%%%%%%%%%%%%%%%%%%%%%%%%%%%%%%%%%%%%%%%%%%%%%%%%%%%%%%%%%%%%%%%%%%%%%



% \input{acknowledgement}

\balance

% References

\bibliographystyle{IEEEtran}
\bibliography{reference}

\end{document}

%% file: abstract.tex
This paper presents a LiDAR odometry estimation framework called Generalized LOAM.
Our proposed method is generalized in that it can seamlessly fuse various local geometric shapes around points to improve the position estimation accuracy compared to the conventional LiDAR odometry and mapping (LOAM) method.
To utilize continuous geometric features for LiDAR odometry estimation, we incorporate tiny neural networks into a generalized iterative closest point (GICP) algorithm.
These neural networks improve the data association metric and the matching cost function using local geometric features.
Experiments with the KITTI benchmark demonstrate that our proposed method reduces relative trajectory errors compared to the other LiDAR odometry estimation methods.

%% file: introduction.tex
\section{INTRODUCTION}
\label{sec:introduction}
\IEEEPARstart{L}{iDAR} odometry estimation with three-dimensional (3D) point clouds is essential for mobility technologies, such as autonomous mobile robots.
To improve the accuracy and efficiency of the estimates, the use of surface geometric features estimated from point clouds is a promising approach.
Some state-of-the-art LiDAR odometry estimation methods \cite{lego_loam} employ feature-based registration derived from LiDAR odometry and mapping (LOAM) \cite{loam}.
LOAM classifies points into edges and planes before registration.
This approach enables efficient and accurate odometry estimates using local geometric shapes.
However, LOAM has several downsides, such as the need for sensor-specific processing, difficulty in adjusting the balance between edges and planes, and degeneracy in environments where features cannot be well extracted.

Generalized iterative closest point (GICP) \cite{gicp} is another popular method.
GICP models the local surface at each point as a normal distribution and incorporates normal distribution comparison into the ICP matching cost.
In the GICP framework, the sum of the distribution-to-distribution distances between corresponding points is minimized instead of the conventional point-to-point distance. 
Because a normal distribution can represent various local geometric shapes such as spheres, lines, or planes depending on the covariance matrix, GICP can be considered an ICP for a local-shape-to-local-shape comparison.
The GICP cost function encompasses the edge and plane comparisons used in LOAM depending on the covariance matrices, and it can handle them seamlessly without explicitly separating the point cloud.

However, in the standard GICP, the normal distributions representing local geometric shapes around points are regularized as planes, which turns GICP into a plane-to-plane ICP \cite{gicp}.
This regularization is necessary to deal with rank deficient of covariance matrices, regardless of the density of the points or sensor noise. 
Unfortunately, this approximated representation leads to inaccurate representations of the local shapes and poorer registration performance.
In addition, there is also room for improvement not only in the cost function, but also in the data association method, for example, in the metric-based ICP approach \cite{metric_icp}.

In this work, we propose a method called Generalized LOAM that fuses various local geometric shapes around points to improve the position estimation accuracy.
Our proposed method is generalized in that it includes the line and plane comparison used in LOAM and fuses various surface shape features to assist in LiDAR odometry estimation.
We reformulate a portion of the functions in the \textit{classical} GICP method with neural networks to be able to seamlessly handle geometric features.
First, we extract intermediate feature vectors using neural networks, such as PointNet++ \cite{pointnet++} and RandLA-Net \cite{randlanet}, which are trained for semantic segmentation.
The feature vectors contain rich information on the local geometric shapes around points.
Then, we improve the following two points in the GICP algorithm with intermediate feature vectors:
\begin{enumerate}
    \item Reducing incorrect data associations by considering the features of points in addition to the Euclidean distance for nearest neighbor search. 
    \item Estimating reasonable covariance matrices for points with feature vectors. The covariance matrices represent local geometric shapes in the GICP matching cost. By improving the covariance estimation, we aim to make point comparisons robust in environments where plane-to-plane ICP estimation accuracy can deteriorate.
\end{enumerate}
For each of data association and covariance estimation, we train tiny multi-layer perceptrons (MLPs) to transform the intermediate feature vectors of semantic segmentation networks into features dedicated to the specific tasks.
Finally, through evaluation on the KITTI dataset \cite{kitti}, we demonstrated that the proposed framework improves the estimation accuracy compared to baselines.

In summary, the key contributions of this work are threefold:
\begin{enumerate}
    \item We present a LiDAR odometry estimation framework called Generalized LOAM. 
    This framework improves data association and covariance matrix estimation in the GICP method. 
    We propose two tiny MLPs that convert local geometric feature vectors to suitable values for data association and matching cost evaluation with GICP. 
    \item We show that Generalized LOAM increases the accuracy of the estimated pose compared to the other existing methods.
    An ablation study confirmed the contribution of the two modified functions, data association, and covariance matrix estimation, to improve the estimation performance.  
\end{enumerate}

%% file: related_works.tex
\section{RELATED WORK}
\label{sec:related_works}
\subsection{Improvement of ICP-based Algorithm Based on Features} 
Because our proposed method is an extension of the GICP algorithm \cite{gicp}, we describe improvements of the ICP-based method based on local features in detail.

There have been several studies on improving data association algorithms using local features.
GF-ICP \cite{gf_icp} improves the cost function for data association and rejects incorrect correspondences using the curvature, surface normal, and density of points.
Gressin {\it et al.} used similar low-level local geometric features to find the optimal radius for the nearest neighbor search \cite{gressin2013towards}.
Color-GICP \cite{color_gicp} finds the nearest neighbors in the extended dimensional space based on the L*a*b* color space.

Some studies proposed methods to estimate the normal distributions of points to better represent local geometric shapes.
Surface-based GICP \cite{surface_based_gicp} estimates the surface normals for shapes using polynomial function fitting and weighted covariance matrices.
To deal with the inhomogeneity of the point cloud, Mesh-GICP \cite{mesh_gicp} estimates the direction of normal distributions from topological information with surface reconstruction.
Multi-channel GICP \cite{multi_chanel_gicp} incorporates additional information, such as color and intensity, into the nearest neighbor search as kernel weights for estimating covariance matrices.
These methods assume that all points belong to planes, which results in extracting low-dimensional local features.

\subsection{Semantic Registration}
Semantic registration methods extract geometric features, such as lines, planes, and spheres, from points, and efficiently determine the correspondences.
LOAM \cite{loam} is the first method that divides point clouds into edge and plane points based on local curvature in a LiDAR scan line.
While this method enables fast point cloud registration, its accuracy and robustness depend significantly on accurate point type labeling.

To achieve a more accurate and robust estimation, several modifications for semantic registration have been developed, including taking into account the uncertainty of the labels \cite{semantic_slam_prob, semantic_slam_akai}, using multi-class semantic segmentation \cite{suma++, mulls, se_ndt_2, semantic_slam_cnn}, and adaptively changing the matching costs for the planar and line points \cite{plane_edge_slam}.
However, these approaches cannot take full advantage of the geometric features of a shape because they classify points into discrete labels and thus cannot seamlessly fuse the data into continuous geometric features.

\subsection{Learning-based Odometry Estimation Approach}
Several works have applied recent machine learning advances to odometry estimation.
One type of method involves direct pose regression using end-to-end LiDAR odometry estimation neural networks \cite{velas2018cnn, lo_net, cho2020unsupervised}.
These approaches do not require determining point-to-point correspondence, unlike the ICP-based method, and thus enable fast registration. 
However, their performance is typically limited to the dataset used for training, and their explainability is low.

Several other methods reformulate a subset of functions in the classical odometry estimation system with neural networks to achieve robust and explainable estimation \cite{back_to_the_future, gradslam, ba_net}.
These methods incorporate trainable features into a conventional optimization framework.
Our proposed method also falls into this category.

%% file: GeneralizedLOAM.tex
\section{GENERALIZED LOAM}

\subsection{System Overview}
\label{sec:system_overview}
\begin{figure*}[t]
  \centering
  \includegraphics[width=0.8\linewidth]{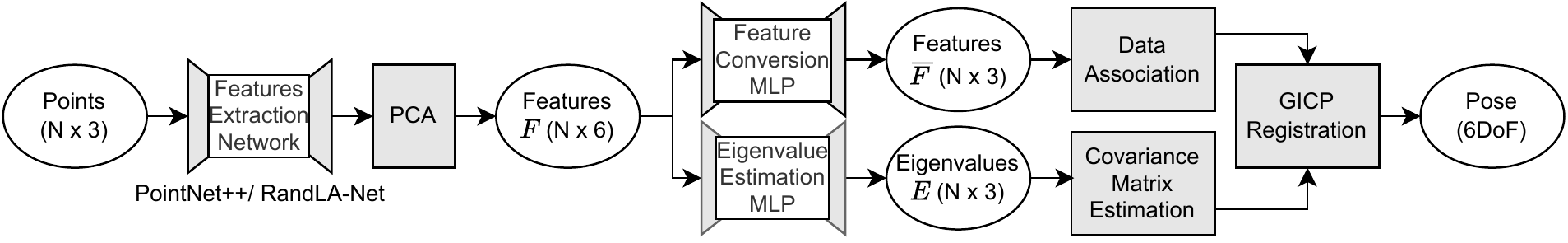}
	\caption{Configuration of the proposed system. 
	The proposed system estimates 6DoF poses on the GICP framework. We improve Data Association and Covariance Matrix Estimation using local geometric features extracted from the pre-trained backbone neural network.}
  \vspace{-2mm}
  \label{fig:system_overview}
\end{figure*}

We aim to improve the following two points in the GICP algorithm with local geometric features:
\begin{enumerate}
    \item Reducing incorrect point correspondences by considering local geometric features.
    \item Providing reasonable point covariance matrices for robust point comparison in environments where plane-to-plane ICP estimation accuracy deteriorates. 
\end{enumerate}
Figure \ref{fig:system_overview} shows the configuration of the proposed system. 
We first use a pre-trained neural network to estimate the local geometric features of points. 
We then reduce the dimensions of the features by principal component analysis (PCA) and transform them into representations dedicated to data association and covariance matrix regularization using tiny MLPs. Then, the pose is estimated following the GICP registration procedure. 

\subsection{GICP Registration}
\label{sec:gicp}
Because our proposed method is an extension of the GICP algorithm \cite{gicp}, we first introduce the details of this algorithm.
We consider the estimation of a pose transformation $\myarray{T}$, which aligns a set of points $\myset{A}=\{a_1, \dots a_{N_A}\}$ (source point cloud) to another set of points $\myset{B}=\{b_1, \dots b_{N_B}\}$ (target point cloud).
Following the conventional ICP scheme, GICP estimates the transformation between point clouds by repeating a two-step process: 1) estimating data association and 2) pose optimization by minimizing a cost function.

In the 1) data association step, the correspondences between source $\myset{A}$ and target $\myset{B}$ point clouds are estimated by the nearest neighbor search:
\begin{align}
    & \{ (a_i, b_i) \; | \; b_i = \argmin_{j \in \{1, \cdots, N_B\}}   \|\myvec{p}_i^a - \myvec{p}_i^b\|^2 \},  \; \forall i \in \{1, \cdots, N_A\}, \label{eq:data_association_gicp} 
\end{align}
where $\myvec{p}_*$ are the Euclidean coordinates of points.

The GICP algorithm models the local surface around a point as a Gaussian distribution: $a_i \sim \mathcal{N}(\hat{a}_i, \myarray{C}_i^\myset{A})$, $b_i \sim \mathcal{N}(\hat{b}_i, \myarray{C}_i^\myset{B})$;
$\myarray{C}_i^\myset{A}$ and $\myarray{C}_i^\myset{B}$ are the covariance matrices for points $a_i$ and $b_i$, respectively.
We assume that the transformation error becomes zero under the correct transformation; 
$\hat{d}_i = \hat{b}_i - \myarray{T} \hat{a}_i = 0$.
Then, the distribution of $d_i$ is given by the reproductive property of the Gaussian distribution as
\begin{align}
     {d}_i & \sim \mathcal{N}(\hat{b}_i - \myarray{T}\hat{a}_i, \myarray{C}_i^\myset{B} + \myarray{T}\myarray{C}_i^\myset{A}\myarray{T}^T) = \mathcal{N}(0, \myarray{C}_i^\myset{B} + \myarray{T}\myarray{C}_i^\myset{A}\myarray{T}^T).\label{eq:dist_gicp}
\end{align}

In the 2) transformation optimization step, the pose transformation $\hat{\myarray{T}}$ is updated such that the log-likelihood of (\ref{eq:dist_gicp}) is maximized:
\begin{align}
    \hat{\myarray{T}} & = \argmax_{\myarray{T}} \sum_i \log(p({d}_i))\\ 
    &= \argmin_{\myarray{T}} \sum_i d_i^T(\myarray{C}_i^\myset{B} + \myarray{T}\myarray{C}_i^\myset{A}\myarray{T}^T)^{-1} d_i. \label{eq:const_func_gicp}
\end{align}
This cost function is the sum of the Mahalanobis distances between corresponding points. It can be regarded as the weighting of the distances between the points by the covariance matrices $\myarray{C}_i^\myset{A}$ and $\myarray{C}_i^\myset{B}$. 
The covariance matrix for each point is typically estimated from its $k$ neighbors and represents the local shape around the point.

For example, suppose the eigenvalues of the covariance matrix are $(a, \epsilon, \epsilon)$, $(a, b, \epsilon)$, and $(a, b, c)$ where $\epsilon$ is a sufficiently small value. In this case, they represent a line, a plane, and a sphere, respectively. 
Thus, GICP can be considered as an ICP for local-shape-to-local-shape comparison.

However, in the original work on GICP \cite{gicp}, the covariance matrices are regularized by replacing their eigenvalues with $(1, 1, \epsilon)$. 
This regularization turns GICP into a plane-to-plane ICP. 
This approximation can lead to incorrect or inadequate constraints, which cause pose estimation accuracy to deteriorate.

\subsection{Estimation of Local Geometric Features}
\label{sec:feature_extraction}

\begin{figure}[t]
  \begin{minipage}[t]{\linewidth}
    \centering
    \includegraphics[width=0.8\linewidth]{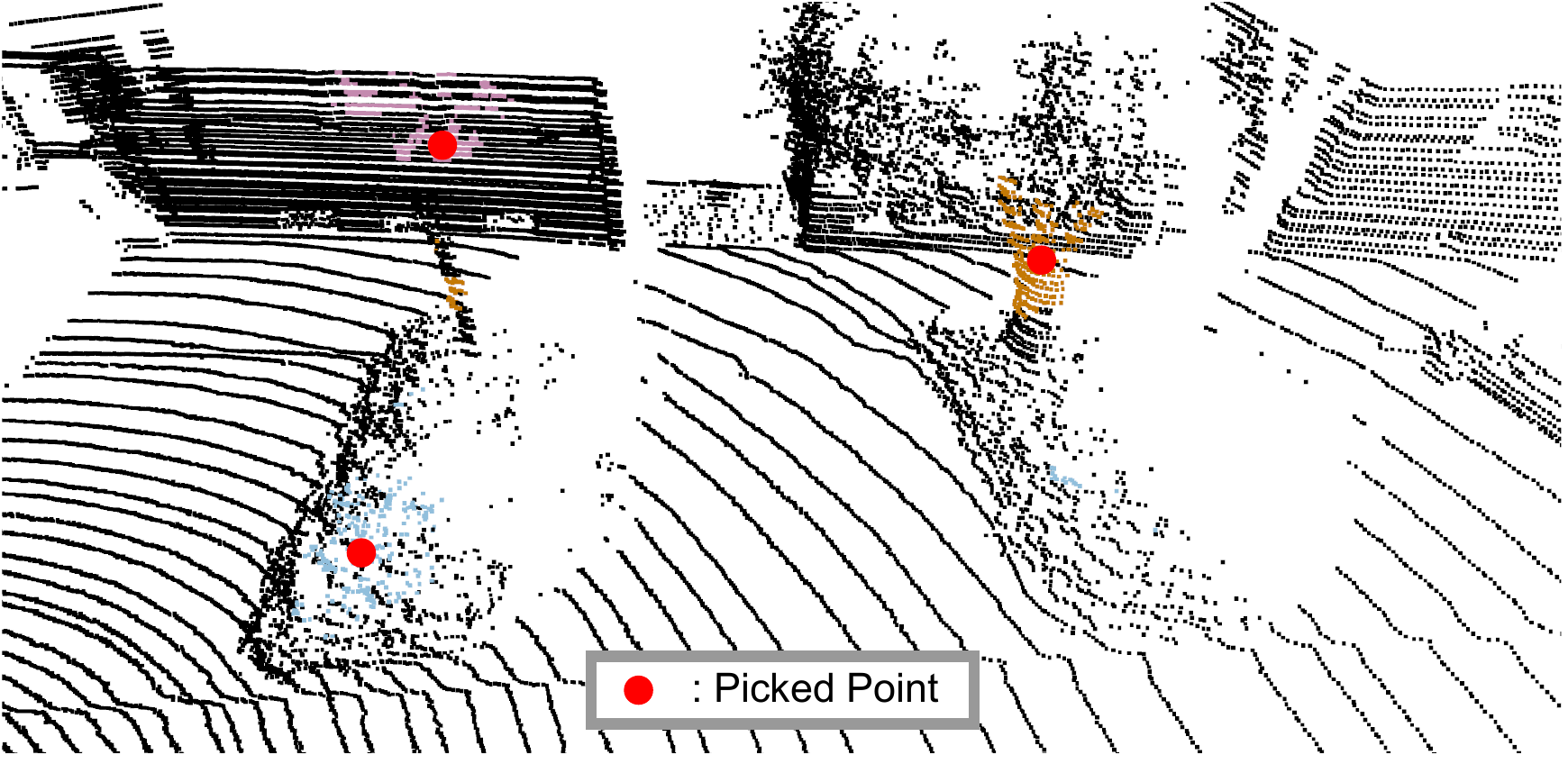}
    \subcaption{Coloring of same semantic labels}
    \label{fig:color_semantic_labels}
  \end{minipage}\\
  \begin{minipage}[t]{\linewidth}
    \centering
    \includegraphics[width=0.8\linewidth]{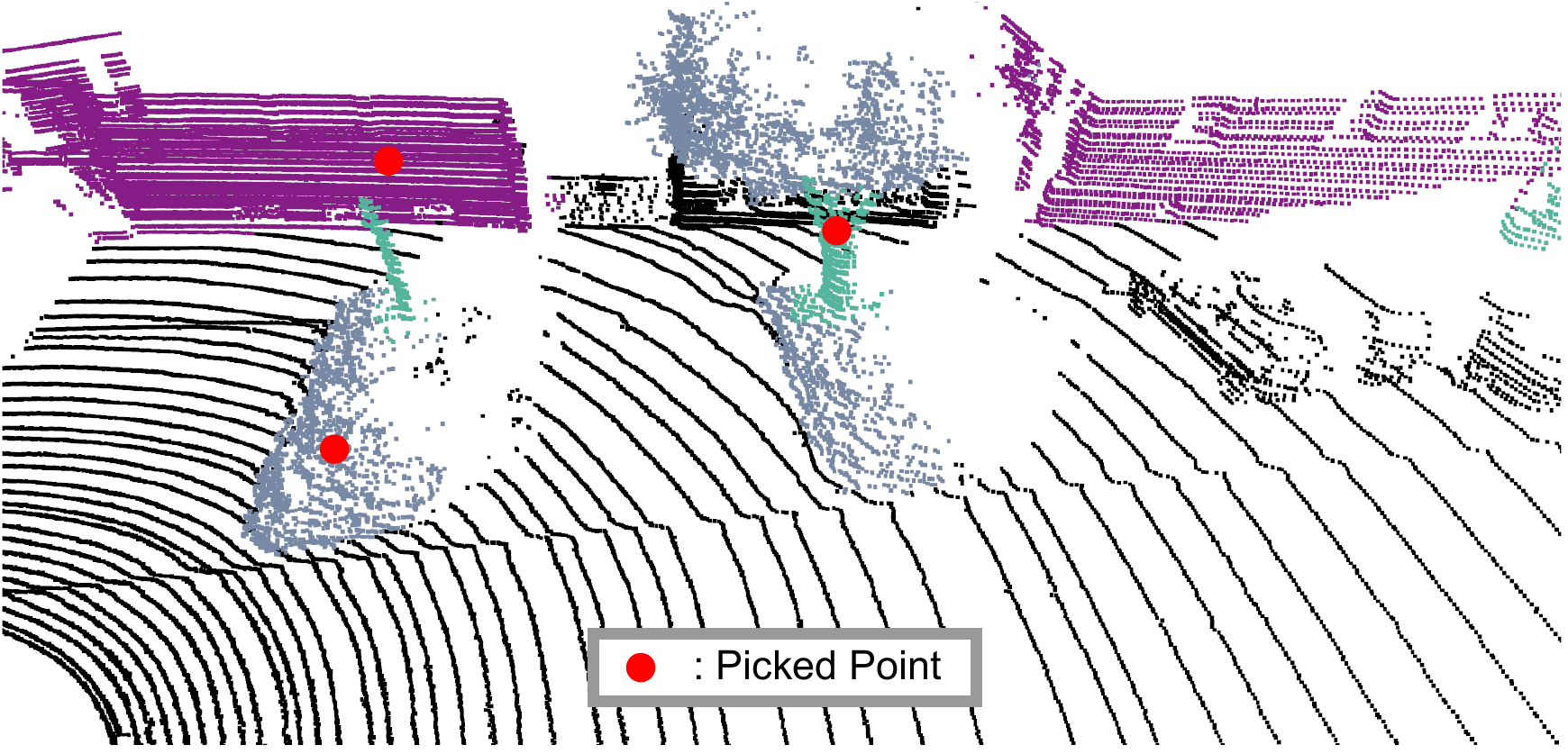}
    \subcaption{Coloring of the close norm of feature vectors}
    \label{fig:color_feature_vectors}
  \end{minipage}
  \caption{Semantic labels (a) and feature vectors (b) obtained from RandLA-Net \cite{randlanet}. 
For each picked point (red circles), we find the 500 closest neighboring points in the feature space and visualize them in the same color. In (b), the colored points surrounding the selected ones have the same labels or are similar in shape and are near in the distance. This result shows that the feature vectors contain information about local geometric shapes and spatial distances.
}
 % \vspace{-3mm}
  \label{fig:feature_vector}
\end{figure}

We estimate local geometric features using a pre-trained neural network for 3D point clouds.
This work uses and compares Pointnet++ \cite{pointnet++} and RandLA-Net \cite{randlanet}, which are designed and trained for semantic segmentation.
Because the outputs of these networks are the semantic labels of points, we expect their intermediate feature representations to contain rich information on the local geometric shapes around points. 
We extract the intermediate feature vectors from the 4th layer of PointNet++ and the 12th layer of RandLA-Net. 
The sizes of the PointNet++ and RandLA-Net feature vectors are $(N, 128)$ and $(N, 32)$, respectively, where $N$ is the number of points.
These features are used in a later process after they are compressed down to six-dimensional feature vectors $\myset{F} = \{\myvec{f}_1 \dots \myvec{f}_{N}\}$ with PCA-based on singular value decomposition implemented in \cite{scikit-learn}.
The reason for the compressioin is that the extracted raw feature vectors are sparse and the trained parameters need to be reduced, as described in section \ref{sec:conclusion}.
In our experiments on the KITTI dataset \cite{kitti}, 
the compressed features of PointNet++ and RandLA-Net have cumulative contribution rates of over 80\% and 90\%, respectively.

Figure \ref{fig:feature_vector} shows the semantic labels and feature vectors obtained from RandLA-Net. 
We find 500 nearest neighbor points in the feature space for each selected point (red circles) and visualize them in the same color.
In Fig. \ref{fig:feature_vector}, we can see that the colored points around the picked ones have the same labels or are similar in shape and are close in the distance.
This result suggests that the feature vectors contain local geometric shapes and spatial distance information.

\subsection{Data Association Using Local Geometric Features}
\label{sec:data_association}

\begin{figure*}[t]
  \centering
  \includegraphics[width=0.9\linewidth]{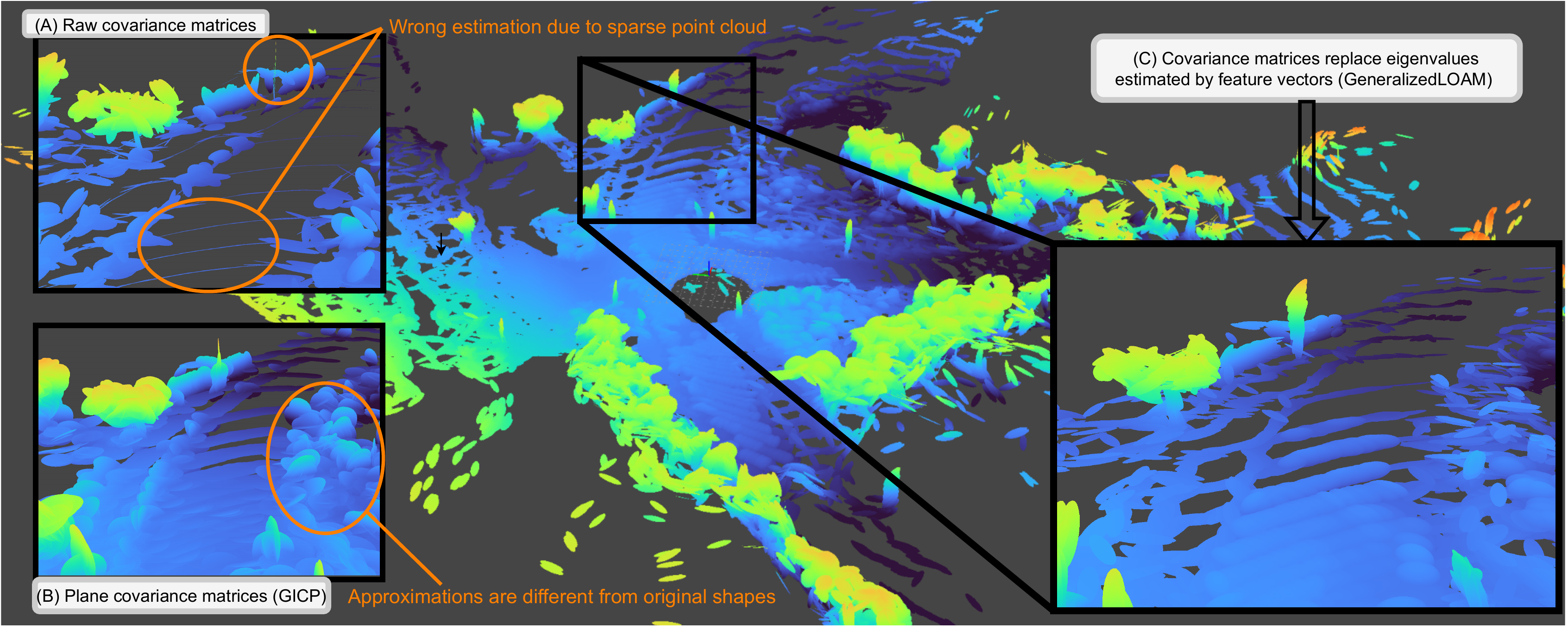}
	\caption{Example of estimated covariance matrix shapes. (A) shows the raw covariance matrices estimated using only the positions of points. 
We can see that some points have degenerate distributions caused by the sparsity of pole and road surface points.
The raw covariance matrices are regularized as planes in the standard GICP method.
(B) shows the plane covariance matrices used in the standard GICP method.
While this approach avoids degenerate distributions, the distributions of some points do not sufficiently represent the original surface shape.
On the other hand, as shown in (C), the shapes of covariance matrices estimated by the proposed regularization algorithm better represent the surface shapes (planes, spheres, and lines) while avoiding the degeneracy of the distributions.}
\vspace{-2mm}
  \label{fig:covariance}
\end{figure*}

Equation (\ref{eq:data_association_gicp}) in the standard GICP method states that the correspondences between source and target points are estimated by a nearest neighbor search based on the spatial distance of points.
This metric is simple, and thus significant initial transformation errors and sensor noise can lead to incorrect associations and sub-optimal local solutions.

To reduce incorrect associations under transformation noise, we concatenate the local geometric features and the Euclidean coordinates of each point and find the nearest neighbors in the extended dimensional space. 
This is a similar approach to \cite{color_gicp}.
We use the L2 norm to measure the distance between feature vectors so that an efficient kd-tree algorithm \cite{kdtree} can be used for the nearest neighbor search:
\begin{align}
    &  \{ (a_i, b_i) \; | \; b_i = \argmin_{j \in \{1, \cdots, N_B\}}  ( \|\myvec{p}^a_i - \myvec{p}^b_i\|^2 + \|\bar{\myvec{f}}_i^a - \bar{\myvec{f}}^b_i\|^2) \}, \label{eq:data_association_ggicp} \\
    & \bar{\myvec{f}}_i = \text{FeatureConversionMLP}(\myvec{f}_i),\; \forall i \in \{1, \dots N_A\},
\end{align}
where $\myvec{p}_*$ is the 3D coordinate of the point and $\myvec{f}_* \in \myset{F}$ is the estimated local geometric feature vector described in Section \ref{sec:feature_extraction}.
Note that positions and features are different physical quantities, and the 6-dimensional feature vector $\myvec{f} \in \myset{F}$ should not be used as-is.
In \cite{color_gicp}, these scales are adjusted using weight parameters, but a trial-and-error adjustment is required.
Instead of introducing weight parameters, we add a feature encoder (\textit{Feature Conversion MLP}) before the data association, as shown in Fig. \ref{fig:system_overview}.
This encoder is trained to convert the feature vector $\myvec{f} \in \myset{F}$ to an appropriate vector $\myvec{\bar{f}} \in \myset{\bar{F}}$ for the data association in Eq. (\ref{eq:data_association_ggicp}). 
The encoder is a three-layer MLP consisting of six, four, and three nodes in the input, middle, and output layers, respectively. The ReLU activation function is used. 
The number of dimensions for output feature vector $\bar{\myvec{f}} \in \myset{\bar{F}}$ used in Eq. (\ref{eq:data_association_ggicp}) is three.
The training procedure for the \textit{Feature Conversion MLP} will be described in section \ref{sec:learn}.

\subsection{Covariance Matrix Estimation Using Local Geometric Features}
\label{sec:covariance_estimation}

\begin{algorithm}[t]
  \caption{Covariance Matrix Estimation}
  \label{alg:cov_estimate}
  \begin{algorithmic}[1]
    \Require{Point coordinates $\myvec{p}_i \in \myset{P}$, Point features $\myvec{f}_i \in \myset{\bar{F}}$, Number of neighbors $k$, Eigenvalue threshold $\epsilon$}
    \Ensure{Estimated covariance matrix $\myarray{C}_i$}
    \State{$\myset{M}_i = {\rm KNN}(k, \myvec{p}_i, \myset{P})$}
      \Comment{K-nearest neighbor search}
      \State{$\myvec{\mu}_i = \frac{\sum_{\myvec{p}_j \in \myset{M}_i} \myvec{p}_j}{k}$}
      \State{$\myarray{C}_i = \frac{\sum_{\myvec{p}_j \in \myset{M}_i} \myvec{p}_j \myvec{p}_j^T}{k} - \myvec{\mu}_i \myvec{\mu}_i^T$} \label{alg_eq:cov}
      \State{$\myarray{Q}_i {\Lambda}_i \myarray{Q}_i^{-1} = \myarray{C}_i$}
      \Comment{Eigenvalue decomposition}
      \State{$\myvec{e}_i = \text{EigenvalueEstimationMLP}(\myvec{f}_i)$}
      \State{$\myvec{e}'_i = (\text{Sort $\myvec{e}_i$ elements in ascending order})$}
      \For{$e \in \myvec{e}'_i = \{ e_1, e_2, e_3 \}$}
      \If{$e \leq \epsilon$}
         \State{$e = \epsilon$}
         \Comment{To prevent rank drop}
      \EndIf
      \EndFor
      \State{$\myvec{e}''_i = \myvec{e}'_i / \|\myvec{e}'_i\|_2$}
      \Comment{Scaling eigenvalues}
      \State{$\myarray{C}'_i = \myarray{Q}_i \text{diag}(\myvec{e}''_i) \myarray{Q}_i^{-1}$}
      \Comment{Replacing eigenvalues}
    \State{\Return{$\myarray{C}'_i$}}
  \end{algorithmic}
  \end{algorithm}

The covariance matrices in Eq. (\ref{eq:const_func_gicp}) represent the local shapes around points (line, plane, and sphere); 
their eigenvectors and eigenvalues respectively represent the basis vector directions and lengths for point distributions.
As described in section \ref{sec:gicp}, the standard GICP method replaces eigenvalues with $(1, 1, \epsilon)$;
All local shapes are approximated as planar with the same spread. 

The approximation enables robust estimation of the geometric shapes regardless of the density of the points or sensor noise. 
However, the approximation can degrade the registration performance due to inaccurate representation of the local shapes.

Here, we propose a covariance regularization method with a tiny MLP to keep the expressive capability of covariance matrices while avoiding rank deficiency. 
Algorithm \ref{alg:cov_estimate} describes the proposed covariance regularization method.

There are two key points in this algorithm:
\begin{enumerate}
    \item Estimating principal component directions (i.e., eigenvectors) for a distribution based on the positions of the nearest neighboring points. 
    \item Estimating eigenvalues that control the shape of the distribution with feature vectors $\myset{F}$.
\end{enumerate}
The eigenvalues are estimated by an encoder called \textit{Eigenvalue Estimation MLP}, as shown in Fig. \ref{fig:system_overview}. 
This encoder is a three-layer MLP consisting of six, four, and three nodes in the input, middle, and output layers. The ReLU activation function is used to inject non-linearity.

Because the extracted local geometric features are trained to be rotation-invariant in Euclidean space \cite{pointnet} and do not contain directional information for the local shape, we estimate the eigenvectors with the standard PCA procedure and let the MLP estimate only the eigenvalues.

Figure \ref{fig:covariance} shows a visualization of an example of estimated covariance matrix shapes.
Figure \ref{fig:covariance}(A) shows the raw covariance matrices are estimated using only positions of points (line \ref{alg_eq:cov} in the Algorithm \ref{alg:cov_estimate}). 
We can see that some points have degenerate distributions caused by the sparsity of pole and road surface points.
The raw covariance matrices are regularized as planes in the standard GICP method.
Figure \ref{fig:covariance}(B) shows the plane covariance matrices used in the standard GICP method.
While this approach avoids degenerate distributions, the distributions of some points do not sufficiently represent the original surface shape.
On the other hand, as shown in Fig. \ref{fig:covariance}(C), the shapes of covariance matrices estimated by the proposed regularization algorithm better represent the surface shapes (not only planes but also spheres and lines) while avoiding degeneracy of the distributions.

\subsection{Training Feature Conversion and Eigenvalue Estimation MLPs}
\label{sec:learn}

\begin{figure*}[t]
  \centering
  \includegraphics[width=0.8\linewidth]{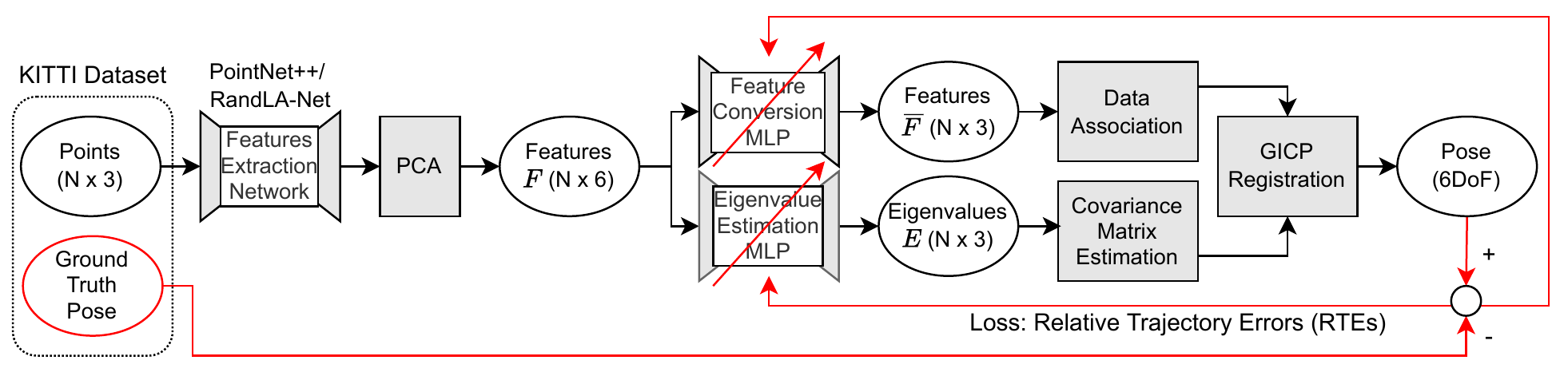}
	\caption{Pipeline for training \textit{Feature Conversion MLP} and \textit{Eigenvalue Estimation MLP} on the KITTI dataset \cite{kitti}. We use the loss function of the translational relative trajectory error (RTE) between the estimated and ground-truth trajectories. The backbones of the \textit{Features Extraction Network} (Pointnet++ \cite{pointnet++} and RandLA-Net \cite{randlanet}) are pre-trained on the SemanticKITTI dataset \cite{semantic_kitti}.}
  \label{fig:how_to_learn}
  \vspace{-2mm}
\end{figure*}

\begin{figure*}[t]
  \centering
  \includegraphics[width=0.9\linewidth]{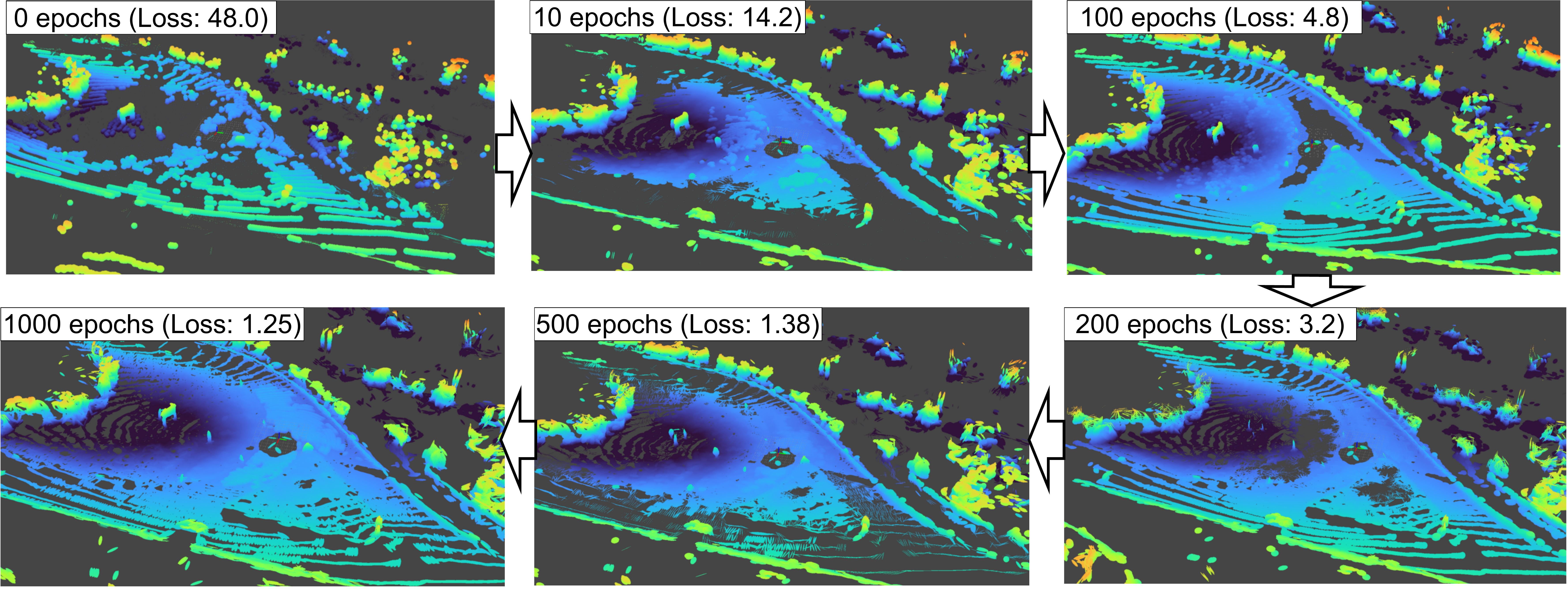}
	\caption{Shapes of the estimated covariance matrix during the training process. In the early stages of training, the shapes of the estimated covariance matrices have many holes and thorns. As the learning progresses, they represent the actual local geometric shapes more closely as the loss decreases. This result suggests that the minimization of the relative pose errors enables the \textit{Eigenvalue Estimation MLP} to represent local geometric shapes.}
  \label{fig:training_process}
  \vspace{-2mm}
\end{figure*}

We train the \textit{Feature Conversion MLP} and \textit{Eigenvalue Estimation MLP} using the KITTI dataset \cite{kitti}.
Figure \ref{fig:how_to_learn} shows the pipeline for training the MLPs. 
Inspired by \cite{back_to_the_future}, we take a closed-loop training approach.
We use the translational relative trajectory error (RTE) \cite{eval_metric} between the estimated and ground-truth trajectories as the loss function $\mathcal{L}$: 

\begin{align}
    & \mathcal{L} = \frac{1}{N}\sum_s \sum_{s \in \myset{T}} \| \delta \myvec{t}_s^{g} - \delta \hat{\myvec{t}}_{s} \|_2,  \label{eq:loss} 
\end{align}
where $s$ represents a sub-trajectory with a fixed length cut from the trajectory $\myset{T}$.
$\delta \myvec{t}_s^{g}$ and $\delta \hat{\myvec{t}}_s$ are relative translational poses from the start point to the endpoint of each sub-trajectory, which are divided from the ground truth and estimated trajectories, respectively. 
$N$ is the number sub-trajectory length settings and is used $N=8$; $(\text{length of } s) \in \{100, 200, 300, 400, 500, 600, 700, 800\}$ [m] following the KITTI official evaluation code (\textit{Development kit}).

Because the GICP registration module in the proposed system is non-differentiable, we apply a black-box optimization method based on the tree-structured Parzen estimator \cite{hyper_parameter_optimization} for optimizing the weights of the MLPs. The implementation is performed in Optuna \cite{optuna}, and the number of weights for each MLP to be optimized is 46.
The backbones of the \textit{Features Extraction Network} (Pointnet++ and RandLA-Net) are pretrained on the SemanticKITTI dataset \cite{semantic_kitti}.

Figure \ref{fig:training_process} shows how the shapes of the estimated covariance matrix changed during the training process.
Before the training (0 epochs), the MLPs with random weights represent a few planar shapes.
As the training progresses from 10 epochs to 200 epochs, the estimated covariances start to better represent the surface as the loss decreases, 
but there is a large hole near the LiDAR.
After training the MLPs for 500 epochs, 
the hole is filled, 
but the shapes of estimated covariance matrices tend to be ``thorny'' due to the sparsity of points and do not represent the actual local geometric shapes well.
However, after 1000 epochs of training, they better represent local surface shapes while avoiding rank deficiency.
The training changes the shapes significantly from the unlearning phase.
This result suggests that minimization of the relative pose errors enables the MLP to represent local geometric shapes.
In other words, it implies that the covariance matrices in the cost function (\ref{eq:const_func_gicp}) can improve the accuracy of the GICP by correctly approximating the local geometric shapes.

%% file: experiments.tex
\section{EXPERIMENT}

\subsection{Implementation Details}

We trained and evaluated the proposed Generalized LOAM method on the KITTI dataset.
We trained the MLPs for 1000 epochs using five sequences (01, 03, 05, 07, 09) out of the 11 sequences of the KITTI dataset and used the rest as a validation set. 
Optimization of the GICP registration module was performed using the Levenberg-Marquardt optimizer in GTSAM\footnote{\url{https://github.com/borglab/gtsam}}.

\subsection{Evaluation}

We calculated the RTE averaged over 100 to 800 m trajectories with the KITTI official evaluation code (\textit{Development kit}). 
We used an Intel Core i9-10900 (16 threads) to run the proposed framework.

\subsubsection{\textbf{Comparison with Baseline Methods}}

% \usepackage{multirow}
\begin{table*}[t]
\centering
\caption{Average translational RTEs [\%] for the KITTI dataset \cite{kitti}}
\label{tab:result}
\begin{threeparttable}
\begin{tabular}{c|cccccc|ccccccc}
\hline
 &
  \multicolumn{6}{c|}{Training dataset} &
  \multicolumn{7}{c}{Test dataset} \\
\multirow{-2}{*}{Seq. ID} &
  01 &
  03 &
  05 &
  07 &
  \multicolumn{1}{c|}{09} &
  Avg. &
  00 &
  02 &
  04 &
  06 &
  08 &
  \multicolumn{1}{c|}{10} &
  Avg. \\ \hline
\rowcolor[HTML]{EFEFEF} 
LOAM &
  {\color{red} 2.33} &
  4.18 &
  5.34 &
  2.02 &
  \multicolumn{1}{c|}{\cellcolor[HTML]{EFEFEF}7.35} &
  4.24 &
  3.51 &
  10.8 &
  3.40 &
  1.19 &
  4.66 &
  \multicolumn{1}{c|}{\cellcolor[HTML]{EFEFEF}4.08} &
  4.61 \\
LeGO-LOAM &
  3.07 &
  1.63 &
  1.02 &
  1.02 &
  \multicolumn{1}{c|}{1.29} &
  1.61 &
  1.87 &
  1.83 &
  1.33 &
  1.09 &
  1.76 &
  \multicolumn{1}{c|}{1.83} &
  1.62 \\
\rowcolor[HTML]{EFEFEF} 
GICP &
  3.18 &
  1.06 &
  0.90 &
  {\color{blue} 0.57} &
  \multicolumn{1}{c|}{\cellcolor[HTML]{EFEFEF}1.26} &
  1.40 &
  1.11 &
  1.54 &
  1.07 &
  0.72 &
  1.15 &
  \multicolumn{1}{c|}{\cellcolor[HTML]{EFEFEF}1.40} &
  1.17 \\
SuMa &
  5.59 &
  1.14 &
  0.87 &
  0.66 &
  \multicolumn{1}{c|}{{\color{red} 0.86}} &
  1.82 &
  {\color{red} 0.84} &
  {\color{red} 1.27} &
  {\color{red} 0.56} &
  {\color{red} 0.60} &
  1.43 &
  \multicolumn{1}{c|}{1.73} &
  {\color{blue} 1.07} \\
\rowcolor[HTML]{EFEFEF} 
Generalized LOAM (PN++) &
  3.16 &
  {\color{blue} 1.04} &
  {\color{blue} 0.82} &
  {\color{red} 0.53} &
  \multicolumn{1}{c|}{\cellcolor[HTML]{EFEFEF}1.15} &
  {\color{blue} 1.34} &
  1.04 &
  1.45 &
  1.01 &
  0.68 &
  {\color{blue} 1.10} &
  \multicolumn{1}{c|}{\cellcolor[HTML]{EFEFEF}{\color{blue} 1.24}} &
  1.09 \\
Generalized LOAM (RN) &
  {\color{blue} 2.91} &
  \cellcolor[HTML]{FFFFFF}{\color{red} 0.95} &
  {\color{red} 0.76} &
  {\color{red} 0.53} &
  \multicolumn{1}{c|}{{\color{blue} 1.00}} &
  {\color{red} 1.23} &
  {\color{blue} 0.94} &
  {\color{blue} 1.33} &
  {\color{blue} 0.87} &
  {\color{blue} 0.64} &
  {\color{red} 1.06} &
  \multicolumn{1}{c|}{{\color{red} 0.99}} &
  {\color{red} 0.97} \\ \hline
\end{tabular}
\begin{tablenotes}[flushleft]
\item[ ] \textcolor{red}{Red} and  \textcolor{blue}{blue} respectively indicate the first and second best results. PN++ and RN mean PointNet++ and RandLA-Net, respectively. LOAM and GICP are implemented by ourselves. LeGO-LOAM and SuMa show the results with parameters tuned by Optuna in \cite{koide2021adaptive}.
\vspace{-3mm}
\end{tablenotes}
\end{threeparttable}
\end{table*}

We compared the proposed method with four LiDAR odometry estimation algorithms: LOAM \cite{loam}, LeGO-LOAM \cite{lego_loam}, plane-to-plane GICP \cite{gicp}, and SuMa \cite{suma}. 
We implemented both LOAM and GICP, and they share common processes (e.g., preprocessing and pose optimization) and parameters for a fair comparison.
The LeGO-LOAM and SuMa results were obtained with parameters tuned by Optuna \cite{optuna} in \cite{koide2021adaptive}.
Note that all algorithms are compared using only the front-end part (loop closure is disabled).
All quantitative evaluation results including rotational RTEs are available on a supplementary website \footnote{See the project page for details: \url{https://kohonda.github.io/proj-gloam/}}.

Table \ref{tab:result} shows the translational RTEs. 
The proposed method using RandLA-Net gives the lowest RTEs ($1.23$\% for the training set and $0.97$\% for the test set).
The proposed method with PointNet++ presents the second-lowest and third-lowest RTEs ($1.34$\% and $1.09$\%, respectively, for the training and test sets).
This result suggests that the proposed method improves the accuracy of GICP due to high-quality local geometric feature representation. 
However, while the proposed method ranks first and second in the training dataset, 
SuMa gives the second-lowest average and lowest RTEs in Seqs. 00, 02, 04, and 06 of the test dataset.
This result indicates that the proposed methods are slightly over-trained.
We expect that the generalization performance can be improved by training on various data other than the KITTI dataset and enhancing the data efficiency of training described in section \ref{sec:conclusion}.

\subsubsection{\textbf{Processing Time}}

\begin{figure}[t]
  \centering
  \includegraphics[width=0.9\linewidth]{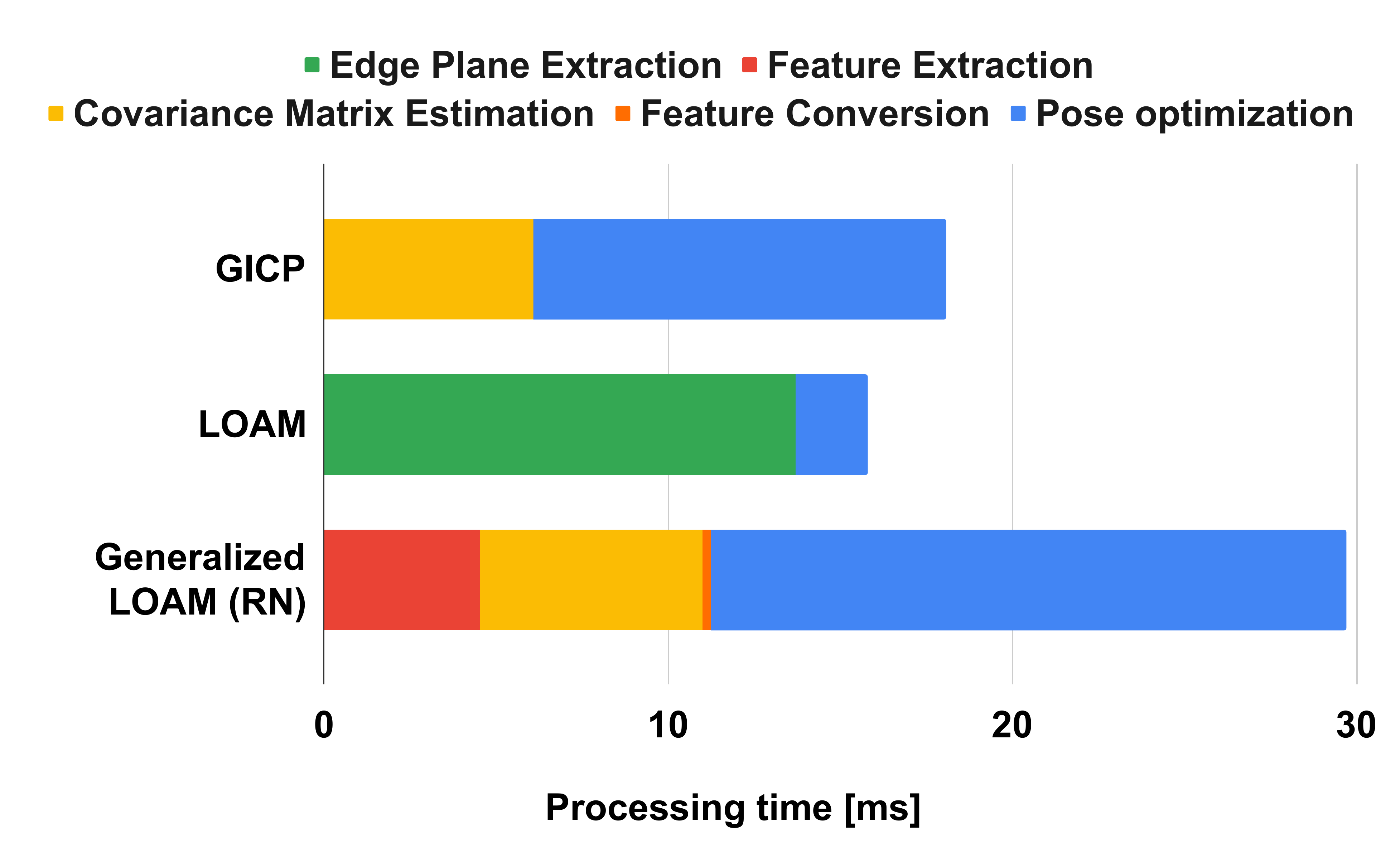}
 \caption{Average processing time per scan. Processing is parallelized with 16 threads. RN means RandLA-Net. Generalized LOAM takes roughly 30 ms to process a frame with multi-threading and thus is sufficiently faster than the real-time requirement (100 ms per frame).}
 \vspace{-3mm}
  \label{fig:processing_time}
\end{figure}

We compared the processing times for each process for all the methods, and the results are shown in Fig. \ref{fig:processing_time}.
Note that the computation time of LOAM is only for the front-end part.
Because feature extraction using PointNet++ takes an average of 261 ms and is not suitable for real-time execution, only the proposed method using RandLA-Net is shown in Fig. \ref{fig:processing_time}.
The proposed method using RandLA-Net takes roughly 30 ms for processing a frame with multi-threading and thus is sufficiently faster than the real-time requirement (100 ms per frame).
The processing times are, however, slightly increased compared to the baseline methods.
This is because the number of dimensions is increased from 3 to 6, which affects the efficiency of the nearest neighbor search for data association.

\subsubsection{\textbf{Ablation Study}}

\begin{table*}[t]
\centering
\caption{Ablation Study on KITTI Dataset \cite{kitti}}
\label{tab:ablation_study}
\begin{threeparttable}
\begin{tabular}{cc|cc|cc|cc|cc|cc|cc|cc}
\hline
\multicolumn{2}{c|}{} & \multicolumn{2}{c|}{00} & \multicolumn{2}{c|}{02} & \multicolumn{2}{c|}{04} & \multicolumn{2}{c|}{06} & \multicolumn{2}{c|}{08} & \multicolumn{2}{c|}{10} & \multicolumn{2}{c}{mean} \\ \cline{3-16} 
\multicolumn{2}{c|}{\multirow{-2}{*}{Method/Seq.}} & $t_{\rm{RTE}}$\tnote{1} & $r_{\rm{RTE}}$\tnote{1} & $t_{\rm{RTE}}$ & $r_{\rm{RTE}}$ & $t_{\rm{RTE}}$ & $r_{\rm{RTE}}$ & $t_{\rm{RTE}}$ & $r_{\rm{RTE}}$ & $t_{\rm{RTE}}$ & $r_{\rm{RTE}}$ & $t_{\rm{RTE}}$ & $r_{\rm{RTE}}$ & $t_{\rm{RTE}}$ & $r_{\rm{RTE}}$ \\ \hline
\multicolumn{2}{c|}{GICP (Baseline)} & 1.11 & 0.52 & 1.54 & 0.55 & 1.07 & 0.67 & 0.72 & 0.36 & 1.15 & 0.47 & 1.40 & 0.61 & 1.27 & 0.52 \\ \hline
\multicolumn{1}{c|}{} & \cellcolor[HTML]{EFEFEF}PN++ \tnote{3} & \cellcolor[HTML]{EFEFEF}1.06 & \cellcolor[HTML]{EFEFEF}0.50 & \cellcolor[HTML]{EFEFEF}1.48 & \cellcolor[HTML]{EFEFEF}0.53 & \cellcolor[HTML]{EFEFEF}1.03 & \cellcolor[HTML]{EFEFEF}0.64 & \cellcolor[HTML]{EFEFEF}0.70 & \cellcolor[HTML]{EFEFEF}\textbf{0.35} & \cellcolor[HTML]{EFEFEF}1.14 & \cellcolor[HTML]{EFEFEF}0.45 & \cellcolor[HTML]{EFEFEF}1.32 & \cellcolor[HTML]{EFEFEF}0.57 & \cellcolor[HTML]{EFEFEF}1.24 & \cellcolor[HTML]{EFEFEF}0.50 \\ \cline{2-16} 
\multicolumn{1}{c|}{\multirow{-2}{*}{GICP+DA \tnote{2}}} & RN \tnote{3} & 1.07 & 0.50 & 1.49 & 0.53 & 1.02 & 0.64 & 0.68 & 0.35 & 1.12 & 0.45 & 1.28 & 0.55 & 1.22 & 0.50 \\ \hline
\multicolumn{1}{c|}{} & \cellcolor[HTML]{EFEFEF}PN++ & \cellcolor[HTML]{EFEFEF}1.08 & \cellcolor[HTML]{EFEFEF}0.51 & \cellcolor[HTML]{EFEFEF}1.52 & \cellcolor[HTML]{EFEFEF}0.54 & \cellcolor[HTML]{EFEFEF}1.06 & \cellcolor[HTML]{EFEFEF}0.65 & \cellcolor[HTML]{EFEFEF}0.70 & \cellcolor[HTML]{EFEFEF}\textbf{0.35} & \cellcolor[HTML]{EFEFEF}1.12 & \cellcolor[HTML]{EFEFEF}0.45 & \cellcolor[HTML]{EFEFEF}1.31 & \cellcolor[HTML]{EFEFEF}0.58 & \cellcolor[HTML]{EFEFEF}1.24 & \cellcolor[HTML]{EFEFEF}0.51 \\ \cline{2-16} 
\multicolumn{1}{c|}{\multirow{-2}{*}{GICP+CE \tnote{2}}{}} & RN & 1.00 & 0.47 & 1.40 & 0.48 & 0.93 & 0.57 & 0.68 & 0.34 & 1.08 & 0.42 & 1.12 & 0.51 & 1.15 & 0.46 \\ \hline
\multicolumn{1}{c|}{} & \cellcolor[HTML]{EFEFEF}PN++ & \cellcolor[HTML]{EFEFEF}\textbf{1.04} & \cellcolor[HTML]{EFEFEF}\textbf{0.49} & \cellcolor[HTML]{EFEFEF}\textbf{1.45} & \cellcolor[HTML]{EFEFEF}\textbf{0.52} & \cellcolor[HTML]{EFEFEF}\textbf{1.01} & \cellcolor[HTML]{EFEFEF}\textbf{0.62} & \cellcolor[HTML]{EFEFEF}\textbf{0.68} & \cellcolor[HTML]{EFEFEF}\textbf{0.35} & \cellcolor[HTML]{EFEFEF}\textbf{1.10} & \cellcolor[HTML]{EFEFEF}\textbf{0.43} & \cellcolor[HTML]{EFEFEF}\textbf{1.24} & \cellcolor[HTML]{EFEFEF}\textbf{0.54} & \cellcolor[HTML]{EFEFEF}\textbf{1.20} & \cellcolor[HTML]{EFEFEF}\textbf{0.48} \\ \cline{2-16} 
\multicolumn{1}{c|}{\multirow{-2}{*}{\begin{tabular}[c]{@{}c@{}}GICP+DA+CE\\ (Generalized LOAM)\end{tabular}}} & RN & \textbf{0.94} & \textbf{0.43} & \textbf{1.33} & \textbf{0.46} & \textbf{0.87} & \textbf{0.53} & \textbf{0.64} & \textbf{0.33} & \textbf{1.06} & \textbf{0.40} & \textbf{0.99} & \textbf{0.45} & \textbf{1.09} & \textbf{0.43} \\ \hline
\end{tabular}
\begin{tablenotes}[flushleft]
\item[1] $t_{\rm{RTE}}$ and $r_{\rm{RTE}}$ are the translation [\%] and rotation [\textdegree /m] RTEs.
\item[2] DA and CE mean data association and covariance matrix estimation using features, which are described in Section \ref{sec:data_association} and \ref{sec:covariance_estimation}, respectively.
\item[3] PN++ and RN mean PointNet++ and RandLA-Net, respectively.
\vspace{-4mm}
\end{tablenotes}
\end{threeparttable}
\end{table*}

Our proposed method (Generalized LOAM) improves two functions of GICP using local geometric features; data association and covariance matrix estimation.
We compared the RTEs for GICP with four settings: the conventional GICP, GICP with the proposed data association, GICP with the proposed covariance matrix estimation, and Generalized LOAM to confirm that both functions contribute to the improvement in accuracy.
We show the results of these ablation studies using the test dataset (Seq. (00, 02, 04, 06, 08, 10)) in Table \ref{tab:ablation_study}.
GICP with two functions (=Generalized LOAM) shows the best result with Pointnet++ and RandLA-Net for all the sequences.
GICP with either function improves the RTEs compared to the baseline GICP (PN++: $-0.07$\%, $-0.04$\textdegree /m; RN: $-0.18$\%, $-0.09$\textdegree /m).
For RandLA-Net, the improvement of the mean RTEs with covariance matrix estimation ($-0.12$\% and $-0.06$\textdegree /m) is more significant than that with data association ($-0.05$\% and $-0.02$\textdegree /m).
This could be the reason that the backbones (PontNet++ and RandLA-Net) are trained for the semantic segmentation task and that the extracted features are better suited for approximating local geometric shapes than data association. 
We expect that the data association could be further improved by re-training the backbone with data association.

\subsection{Real Data Experiment}

\begin{figure*}[t]
  \centering
  \includegraphics[width=0.9\linewidth]{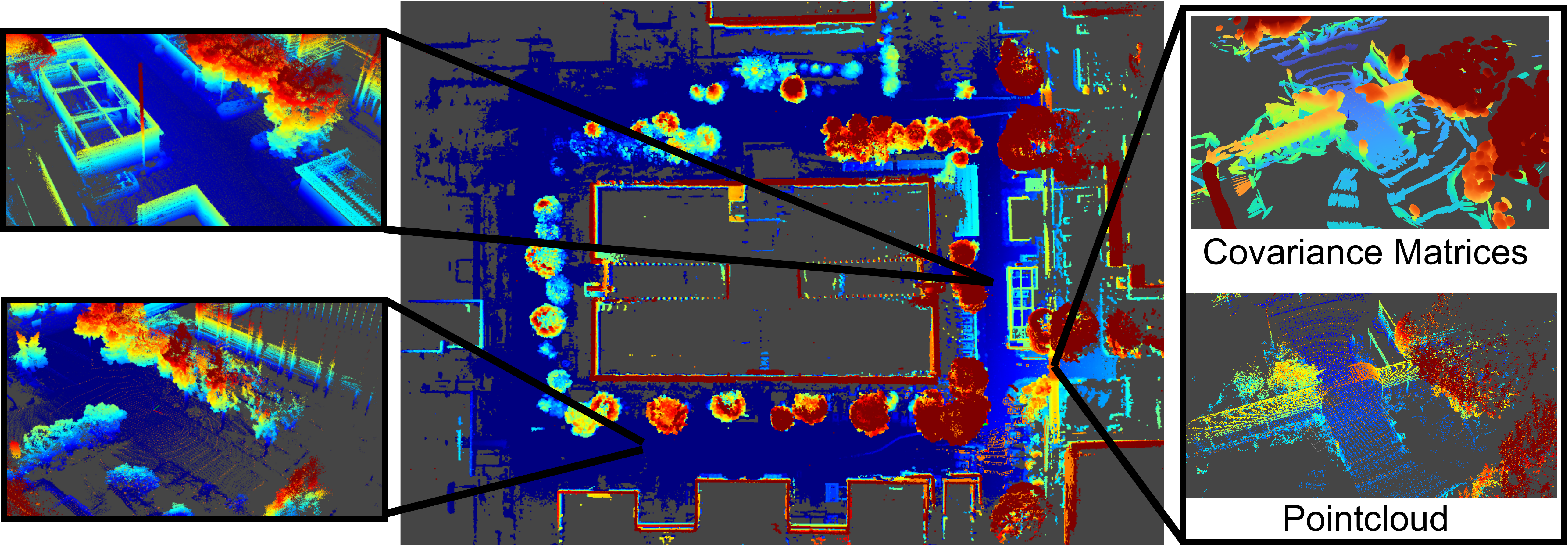}
	\caption{Estimation result on a dataset acquired with Ouster OS1-64. Our proposed system works without major breakdowns and estimates the high-quality local geometric shapes as covariance matrices with a LiDAR model that is different from the one used for the acquisition of the training dataset.}
 % \vspace{-5mm}
  \label{fig:real_data}
\end{figure*}

The Generalized LOAM illustrated in Fig.\ref{fig:system_overview} uses local geometric features extracted from PointNet++ or RandLA-Net as a backbone. 
Because the two neural networks are pre-trained with the KITTI dataset, 
the technical characteristics of the LiDAR model (Velodyne HDL-64E\footnote{\url{https://velodynelidar.com/products/}}) may influence the features.
To verify the generalization performance of the proposed method, 
we tested our system on another dataset acquired with a different LiDAR model.

We measured point cloud data (4142 frames) in an outdoor environment using an Ouster OS0-64 model\footnote{\url{https://ouster.com/jp/products/scanning-lidar/os0-sensor/}}.
Although the number of scan lines in the OS0-64 model is the same as that in Velodyne HDL-64E, the scan angle and range differ.
Figure \ref{fig:real_data} shows a point cloud map and estimated covariance matrices generated by Generalized LOAM with RandLA-Net.
This result shows that the proposed system works well without significant corruption and estimates high-quality local geometric shapes as covariance matrices.

We also compared the pose estimation errors at the start and end frames with the plane-to-plane GICP. The error was 2.56m / 0.023rad for GICP and 1.86m / 0.007rad for Generalized LOAM using RandLA-Net.
This result suggests that the proposed method will perform well when generalized to other LiDAR models.

%% file: conclusion.tex
\section{CONCLUSION AND FUTURE WORK}
\label{sec:conclusion}

This paper presents a point cloud registration method that improves data association and covariance matrix estimation in the GICP method using local geometric features.
We propose two MLPs: the \textit{feature conversion MLP} converts local geometric features to scaled values for the metric for nearest neighbor search, and the \textit{eigenvalues estimation MLP} estimates eigenvalues of the covariance matrices to avoid rank deficient covariance matrices while keeping their expressive capability. These two MLPs are trained with closed-loop pose estimation and an error evaluation pipeline.

In the current implementation, the proposed system involves non-differentiable processes that limit the number of trainable parameters in the system because efficient differentiation-based optimization is not applicable. 
% In future work, we plan to make the entire process differentiable and improve the loss function
% jointly train all the learnable parameters to further improve registration accuracy.
Additionally, although the current loss function in Eq. (\ref{eq:loss}) is a typical and direct evaluation metric for odometry estimation, its results are too sparse to provide teaching data for training; we can obtain only one error for each continuous sequence on the KITTI dataset, which is inefficient for training.
In future work, we plan to further improve registration accuracy by making the entire process differentiable and improving the loss function.